\documentclass{article}
\usepackage{spconf,amsmath,graphicx,hyperref}
\usepackage{enumitem}
\usepackage{amsfonts}
\usepackage{multirow}
\usepackage{booktabs}
\usepackage{makecell}
\usepackage{caption}  
\usepackage{multirow}
\usepackage{booktabs}
\usepackage{upgreek}
\usepackage{siunitx}
\usepackage{xcolor}

\title{CKANIO: Learnable Chebyshev Polynomials for Inertial Odometry}
\name{Shanshan Zhang$^{\star \dagger}$ \qquad Siyue Wang$^{\star}$ \qquad Tianshui Wen$^{\star}$ \qquad Liqin Wu$^{\star}$ \qquad Qi Zhang$^{\star}$ Ziheng Zhou$^{\star}$ 
\qquad Ao Peng$^{\star}$ \qquad Xuemin Hong$^{\star}$ \qquad Lingxiang Zheng$^{\star \ddagger}$ \qquad Yu Yang$^{\dagger \ddagger}$}
\address{$^{\star}$ Department of Information and Communication Engineering, Xiamen University, China \\
         $^{\dagger}$ Department of Electronic Science, Xiamen University, China\\
         $^{\ddagger}$ Corresponding author, lxzheng@xmu.edu.cn, yuyang15@xmu.edu.cn}

\begin{document}
\maketitle
\begin{abstract}
Inertial odometry (IO) relies exclusively on signals from an inertial measurement unit (IMU) for localization and offers a promising avenue for consumer-grade positioning. However, accurate modeling of the nonlinear motion patterns present in IMU signals remains the principal limitation on IO accuracy. To address this challenge, we propose CKANIO, an IO framework that integrates Chebyshev-based Kolmogorov–Arnold Networks (Chebyshev KAN). Specifically, we design a novel residual architecture that leverages the nonlinear approximation capabilities of Chebyshev polynomials within the KAN framework to more effectively model the complex motion characteristics inherent in IMU signals. To the best of our knowledge, this work represents the first application of an interpretable KAN model to IO. Experimental results on five publicly available datasets demonstrate the effectiveness of CKANIO.
\end{abstract}
\begin{keywords}
Chebyshev KAN, Inertial Odometry, Inertial Measurement Unit signals
\end{keywords}

\section{Introduction}
Inertial odometry (IO) estimates the position and orientation of an IMU-equipped platform using acceleration and angular velocity signals provided by the inertial measurement unit (IMU)~\cite{Indoor-Localization-Using-Smartphones}. Because IO does not rely on external cues and can be implemented cost-effectively~\cite{RIO}, it has been widely adopted in autonomous driving and related domains~\cite{CarIMU,surveyILS,SurveyofIndoorInertial}. In scenarios where external sensors (e.g., cameras, radar) or GPS are limited or unavailable, IO serves as a complementary or alternative solution for positioning and navigation~\cite{SCHNN,SSHNN}.

Traditional IO approaches primarily rely on Newtonian mechanics to estimate position and orientation. However, inherent measurement noise in IMUs leads to error accumulation over time, resulting in severe degradation of long-range positioning accuracy~\cite{SINS}. While incorporating physical priors can partially mitigate such errors, this often reduces the generalizability of IO methods and increases system complexity~\cite{Nonlinearity-Aware-ZUPT,AdaptiveThreshold-BasedZUPT}. 

\begin{figure}[t]
\centering
\captionsetup{aboveskip=2pt,font=small}
\includegraphics[width=0.5\textwidth]{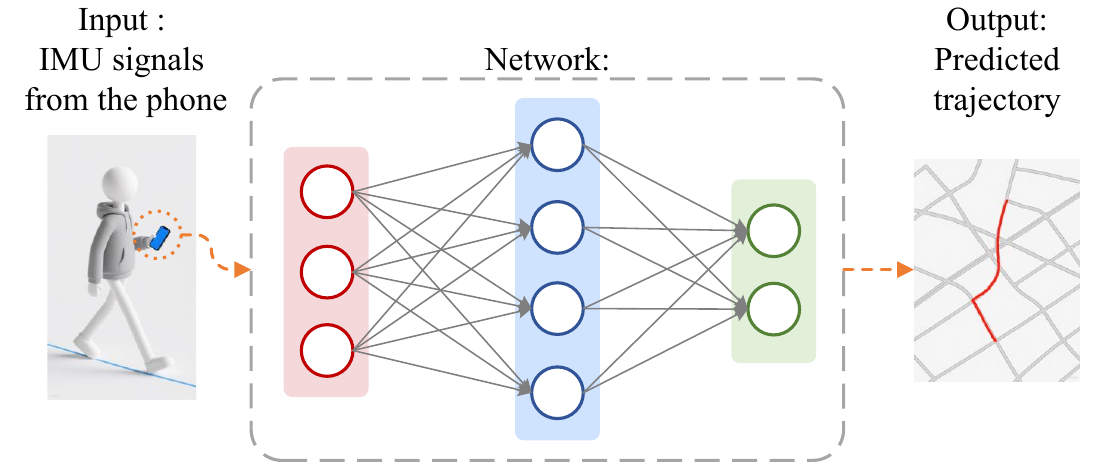}
\caption{Illustration of the processing pipeline for data-driven IO.}
\label{demo}
\vspace{-20pt}
\end{figure}

In contrast, data-driven approaches learn motion patterns and features from large-scale IMU recordings, thereby improving positioning accuracy and broadening the scope of applicable scenarios~\cite{RoNIN}. Their processing pipeline is illustrated in Fig.~\ref{demo}. Nevertheless, these methods may still exhibit significant drift when modeling complex nonlinear motion patterns, such as rotations. This limitation often stems from their reliance on fixed activation functions, which are insufficient to capture the intricate dynamics of IMU signals.

Inspired by recent work on Kolmogorov--Arnold networks (KANs)~\cite{KAN}, several studies have proposed replacing conventional fixed activation functions with polynomial-based alternatives, such as Chebyshev polynomials, to better model complex nonlinear features~\cite{CheyKAN}. Motivated by these developments, we propose CKANIO, which leverages the nonlinear approximation capabilities of a Chebyshev KAN to more effectively model the motion characteristics inherent in IMU signals.

The main contributions of this work are summarized as follows:
\begin{itemize}[noitemsep, nolistsep, leftmargin=*]
    \item We introduce the Chebyshev KAN to IO and design a residual network that exploits its nonlinear approximation capabilities, thereby enhancing the model's representational capacity for nonlinear dynamics.
    \item We develop an efficient kernel-based self-attention module to augment contextual motion modeling and enable more comprehensive utilization of IMU signals.
    \item We present the first quantitative analysis of the impact of gravity on IO accuracy when expressed in the global coordinate frame.
\end{itemize}

\begin{figure}[t] 
\centering
\captionsetup{aboveskip=2pt,font=small} 
\includegraphics[width=0.45\textwidth]{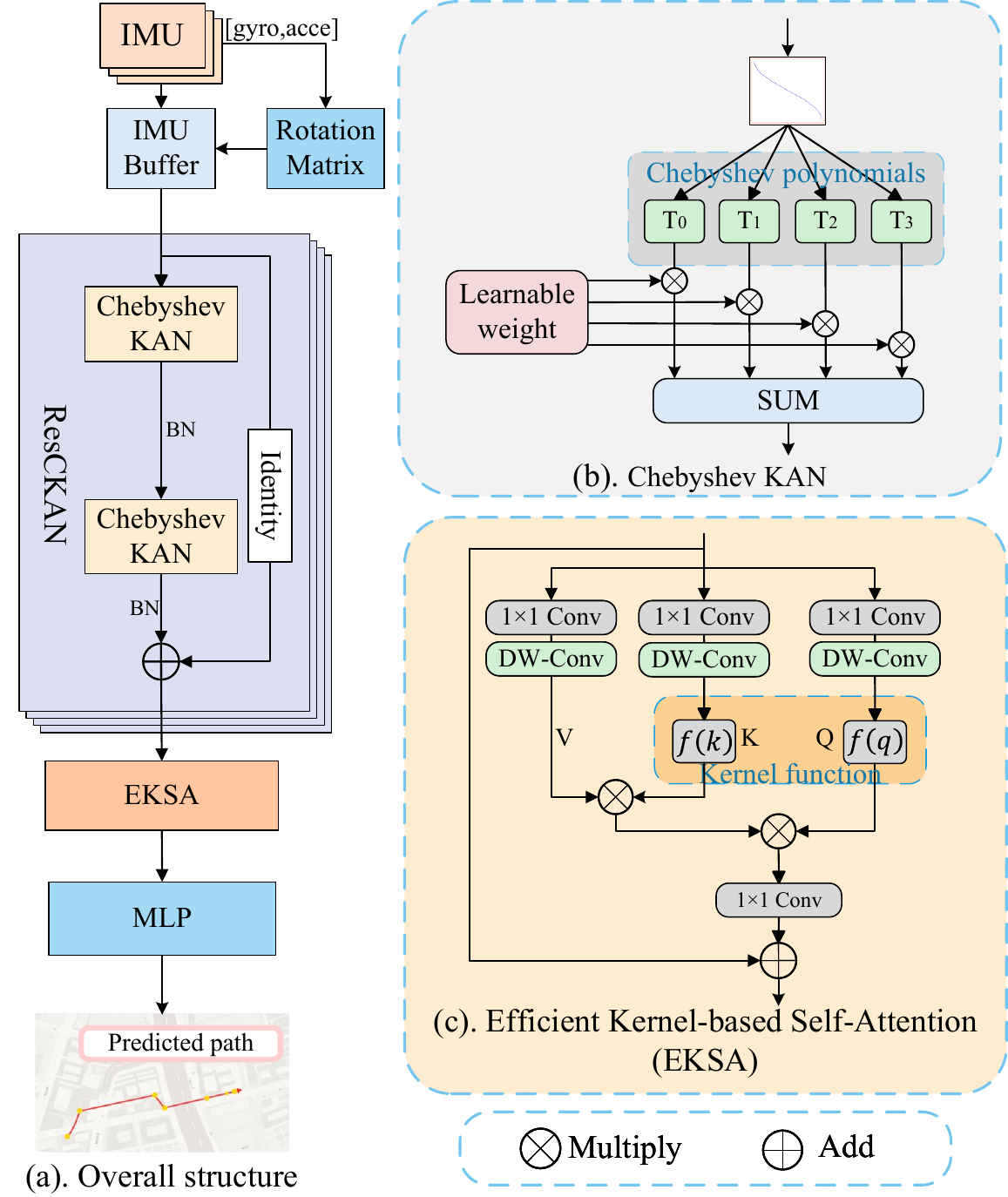}
\caption{Network Architecture Diagram of CKANIO.}
\label{Schematic_Diagram}
\vspace{-15pt} 
\end{figure}

\section{Method}
\subsection{Overall Pipeline}
The overall architecture of CKANIO is depicted in Fig.~\ref{Schematic_Diagram}. CKANIO comprises two primary components: the Residual Chebyshev KAN (ResCKAN) and the Efficient Kernel-based Self-Attention (EKSA) module.

First, we design a one-dimensional (1D) Chebyshev KAN that leverages the strong nonlinear approximation capabilities of Chebyshev polynomials to increase the network's representational power. This one-dimensional Chebyshev KAN forms the core of a residual block—ResCKAN—enabling effective feature extraction with a relatively shallow architecture and promoting stable training dynamics.

Next, we introduce the EKSA module, which employs kernel functions to approximate the computationally expensive attention matrices and reduces computational complexity to linear time with respect to the input length (i.e., \(\mathcal{O}(n)\)).

Because shallow layers predominantly capture high-frequency local features while deeper layers extract global contextual dependencies, we use ResCKAN to first extract nonlinear, high-frequency motion features (e.g., turning motions) that appear locally. The EKSA module then models the deeper contextual dependencies.

Finally, the EKSA output is mapped to a velocity vector \(\hat{\vec{v}}\) via a multilayer perceptron (MLP). The entire CKANIO framework is trained to minimize the mean squared error (MSE) between the predicted velocity \(\hat{\vec{v}}\) and the ground-truth velocity \(\vec{v}\), ensuring accurate velocity estimation.

\subsection{Residual Chebyshev KAN}
Motivated by the success of KANs in various domains\cite{KANconv}, we propose a one-dimensional Chebyshev KAN\cite{CheyKAN}, which represents the first application of a KAN to IO to the best of our knowledge. Rather than using conventional convolutional layers with fixed activation functions, we replace them with a one-dimensional Chebyshev KAN to construct a residual block, termed ResCKAN (see Fig.~\ref{Schematic_Diagram}(a)).

Given IMU signals \(X \in \mathbb{R}^{C \times L}\) collected over a unit time window—where \(C=6\) denotes six channels (three-axis acceleration and three-axis angular velocity) and \(L\) is the number of samples per window—we first apply \(\tanh\) to bound the inputs and then \(\arccos\) to obtain a numerically stable phase representation:
\begin{equation}
X' = \arccos\bigl(\tanh(X)\bigr) \in (0,\pi)^{C \times L}.
\end{equation}
Note that \(\tanh(X)\in(-1,1)\), so \(\arccos(\tanh(X))\) is well-defined; in practice we clip values to \([-1,1]\) if numerical issues arise.

We use Chebyshev polynomials of the first kind, defined for \(x\in[-1,1]\) as
\begin{equation}
T_n(x) = \cos\bigl(n\arccos(x)\bigr).
\end{equation}
Accordingly, for our bounded input we compute
\begin{equation}
T_n\bigl(\tanh(X)\bigr) = \cos\bigl(n\,\arccos(\tanh(X))\bigr),
\end{equation}
where \(n\) denotes the polynomial degree (we set \(n=4\) in our experiments).

To obtain a learnable activation, we form a weighted combination of Chebyshev terms:
\begin{equation}
Y = \sum_{i=0}^{n} W_i \times T_i\bigl(\tanh(X)\bigr) \in \mathbb{R}^{C \times L},
\end{equation}
where \(W_i\) is learnable parameter tensors with shapes compatible with \(T_i(\tanh(X))\). This learnable polynomial activation dynamically adapts to the IMU signals, improving the network's capacity to model complex nonlinear motion dynamics.
The output of ResCKAN is denoted by \(X_{\mathrm{R}} \in \mathbb{R}^{C_R \times L_R}\), where \(C_R\) and \(L_R\) are the output channel dimension and output sequence length, respectively.

\subsection{Efficient Kernel-based Self-Attention}
The input \(X_{\mathrm{R}}^\top\) is linearly projected into queries \(Q\), keys \(K\), and values \(V\), each of shape \(\mathbb{R}^{L_R\times C_R}\). While standard self-attention captures global dependencies, its computational cost scales quadratically with the sequence length—i.e., \(\mathcal{O}(C_R L_R^2)\)—which is prohibitive for long sequences\cite{TKSA}. To address this, we introduce the Efficient Kernel-based Self-Attention (EKSA) module, which approximates the attention matrix using a kernel function.

\begin{table*}[!t]
    \centering
    \captionsetup{aboveskip=2pt,font=small}
    \scriptsize
    \renewcommand{\arraystretch}{1}   
    \caption{
        Cross-dataset performance comparison, including comparison experiments, gravity ablation, and component ablation. 
        Best results in comparison experiments are marked in \textbf{bold}. 
        In gravity ablation (TLIO (w/o gravity)), ↑ and ↓ denote increased and decreased errors relative to TLIO, respectively. 
        Best results in component ablation are marked with \underline{underline}. 
        All metrics are reported in meters.
    }
\label{tab:cross_dataset_algorithm}
    \begin{tabular}{l cc cc cc cc cc |cc| cc l}
        \toprule
        \multirow{2}{*}{\textbf{Method}} & 
        \multicolumn{2}{c}{\textbf{RoNIN}} & 
        \multicolumn{2}{c}{\textbf{RIDI}} & 
        \multicolumn{2}{c}{\textbf{IMUNet}} & 
        \multicolumn{2}{c}{\textbf{RNIN}} & 
        \multicolumn{2}{c}{\textbf{TLIO}} & 
        \multicolumn{2}{c}{\textbf{TLIO (w/o gravity)}} & 
        \multirow{2}{*}{\textbf{Note}} \\
        
        \cmidrule(lr){2-3}  
        \cmidrule(lr){4-5}  
        \cmidrule(lr){6-7}  
        \cmidrule(lr){8-9}  
        \cmidrule(lr){10-11} 
        \cmidrule(lr){12-13} 
        & ATE & RTE & ATE & RTE & ATE & RTE & ATE & RTE & ATE & RTE & ATE & RTE & \\
        \midrule
        RoNIN ResNet     & 5.365 & 3.390 & 2.578 & 2.823 & 14.207 & 7.946 & 1.850 & 2.185 & 1.476 & 3.790 & 1.328↓ & 1.105↓ & \multirow{7}{*}{Comparison} \\
        RoNIN LSTM       & 6.362 & 3.518 & 2.942 & 3.405 & 9.841 & 7.417 & 2.398 & 3.150 & 2.451 & 6.710 & 2.273↓ & 1.684↓ & \\
        RoNIN TCN        & 7.983 & 3.647 & 7.936 & 7.971 & 44.184 & 19.019 & 1.803 & 2.199 & 3.254 & 9.731 & 3.939↑ & 2.215↓ & \\
        RNIN             & 6.987 & 3.438 & 2.170 & 2.296 & 15.174 & 8.164 & 1.432 & 1.776 & 1.572 & 4.183 & 1.371↓ & 1.122↓ & \\
        IMUNet           & 14.591 & 7.967 & 2.909 & 3.335 & 11.847 & 7.792 & 2.005 & 2.631 & 1.432 & \textbf{3.474} & 1.421↓ & 1.116↓ & \\
        TLIO             & 6.481 & 3.746 & 2.332 & 2.524 & 19.958 & 10.123 & 1.553 & 1.865 & 1.493 & 3.628 & 1.645↑ & 1.227↓ & \\
        CKANIO  & \textbf{3.814} & \textbf{3.272} & \textbf{1.835} & \textbf{1.954} & \textbf{9.023} & \textbf{7.082} & \textbf{1.067} & \textbf{1.260} & \textbf{1.420} & 3.509 & \textbf{1.322↓} & \textbf{1.118↓} & \\
        \midrule 
        CKANIO (w/o EKSA) & 5.057 & \underline{3.215} & 2.128 & 2.198 & 9.578 & 7.375 & 1.228 & 1.346 & 1.430 & 3.552 & - & - & \multirow{2}{*}{Ablation} \\
        Full CKANIO       & \underline{3.814} & 3.272 & \underline{1.835} & \underline{1.954} & \underline{9.023} & \underline{7.082} & \underline{1.067} & \underline{1.260} & \underline{1.420} & \underline{3.509} & - & - & \\
        \bottomrule
    \end{tabular}
    \vspace{-15pt}  
\end{table*}

\begin{figure}[t] 
\centering
\captionsetup{aboveskip=2pt,font=small} 
\includegraphics[width=0.5\textwidth]{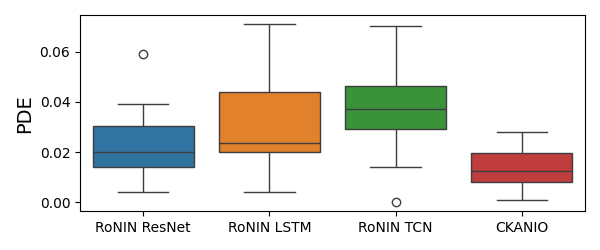}
\caption{PDE comparison between CKANIO and three baseline models from RoNIN on the RoNIN dataset.}
\label{RoNIN}
\vspace{-18pt} 
\end{figure}

To avoid explicitly forming the full attention matrix, we measure the similarity between a query vector \(q\in\mathbb{R}^{C_R}\) and a key vector \(k\in\mathbb{R}^{C_R}\) using the squared Pearson correlation coefficient. Let \(\mu(\cdot)\) denote the mean across vector components and define \(\bar{q}=q-\mu(q)\) and \(\bar{k}=k-\mu(k)\). Then the squared Pearson coefficient is $\uprho(q,k)^2 \;=\; \left(\frac{\bar{q}^\top \bar{k}}{\|\bar{q}\|\,\|\bar{k}\|}\right)^{2}$,
which is invariant to additive offset and multiplicative gain, thereby improving robustness to offset/gain noise commonly observed in IMU signals. The Pearson-attention operation is defined as
\begin{equation}
\text{Pearson-Att}(Q,K,V) \;=\; \uprho(Q,K)^2 \, V \in \mathbb{R}^{L_R\times C_R}.
\end{equation}

To further relieve
the computation burdenn, we construct an exponential kernel based on the squared Pearson similarity:
\begin{equation}
\mathrm{KF}(q,k) \;=\; \exp\!\bigl(\uprho(q,k)^2\bigr).
\end{equation}
Existing research~\cite{ESSA} has shown that \(\mathrm{KF}\) is a Mercer kernel; thus there exists a feature mapping \(f(\cdot)\) such that $\mathrm{KF}(q,k) \;=\; \langle f(q), f(k)\rangle$.
Let \(\tilde{q}=\bar{q}/\|\bar{q}\|\) and \(\tilde{k}=\bar{k}/\|\bar{k}\|\). Using the Taylor expansion of the exponential, we obtain
\begin{equation}
\begin{aligned}
\mathrm{KF}(q,k) &= \exp\!\bigl((\tilde{q} \tilde{k}^\top)^2\bigr)= \sum_{n=0}^{\infty} (\tilde{q} \tilde{k}^\top)^{2n}/n! \\
&= \sum_{n=0}^{\infty} \frac{\tilde{q}^{2n}}{\sqrt{n!}} \cdot \left(\frac{\tilde{k}^{2n}}{\sqrt{n!}}\right)^\top.
\end{aligned}
\end{equation}
A valid feature mapping is therefore
$
f(q)=\sum_{n=0}^{\infty} \frac{\tilde{q}^{2n}}{\sqrt{n!}}.
$
With the finite mapping, EKSA is computed as
\begin{equation}
\mathrm{EKSA}(Q,K,V) \;=\; \bigl(f_Q f_K^\top\bigr) V \;=\; f_Q \bigl(f_K^\top V\bigr),
\end{equation}
Hence, the EKSA reduces the computational complexity from \( \mathcal{O}(C_R L_R^2) \) to \( \mathcal{O}(L_R C_R^2) \). This formulation captures contextual motion information in IMU signals with linear complexity in \(L_R\) and enhances robustness to gain and offset noise.

\section{Experiments}
\subsection{Implementation Details}
\textbf{Datasets and Metrics.}
We use five publicly available inertial datasets: IMUNet\cite{IMUNet}, RoNIN\cite{RoNIN}, RIDI\cite{RIDI}, RNIN\cite{RNIN-VIO}, and TLIO\cite{TLIO}. Each dataset is randomly partitioned into training, validation, and test subsets with an 8:1:1 ratio. For evaluation, we adopt three widely used metrics—Position Drift Error (PDE)\cite{CTIN}, Relative Trajectory Error (RTE)\cite{IDOL}, and Absolute Trajectory Error (ATE)\cite{IDOL}—to quantify terminal drift, local trajectory accuracy, and global positioning accuracy, respectively.

To analyze the effect of gravity in the global coordinate frame, we remove the gravitational acceleration from the TLIO global-coordinate IMU signals and denote the resulting dataset as TLIO (w/o gravity). Models are retrained on both the original TLIO and TLIO (w/o gravity) in order to quantify the influence of gravity on IO performance.

\textbf{Comparison Methods.}
Data-driven IO approaches have shown superior accuracy compared to traditional methods\cite{iMOT,IMU-and-magnetometer-modeling-for-smartphone-based-PDR,AirIO}; thus, we compare CKANIO against classic learning-based IO methods. The baselines include RoNIN (ResNet, TCN, and LSTM variants)\cite{RoNIN}, IMUNet\cite{IMUNet}, and the networks introduced in TLIO\cite{TLIO} and RNIN\cite{RNIN-VIO}.

\textbf{Experimental Settings.}
All training and evaluation are performed on an NVIDIA RTX A40 GPU (48 GB). The initial learning rate is set to \(10^{-4}\), the batch size is 512, and training runs for up to 100 epochs. The CUDA version used is 11.3, and the PyTorch version is 1.11. All other experimental settings were kept consistent with those of TLIO\cite{TLIO}.

\begin{figure*}[!t]
\centering
\captionsetup{aboveskip=2pt,font=small}
\includegraphics[width=1.0\textwidth]{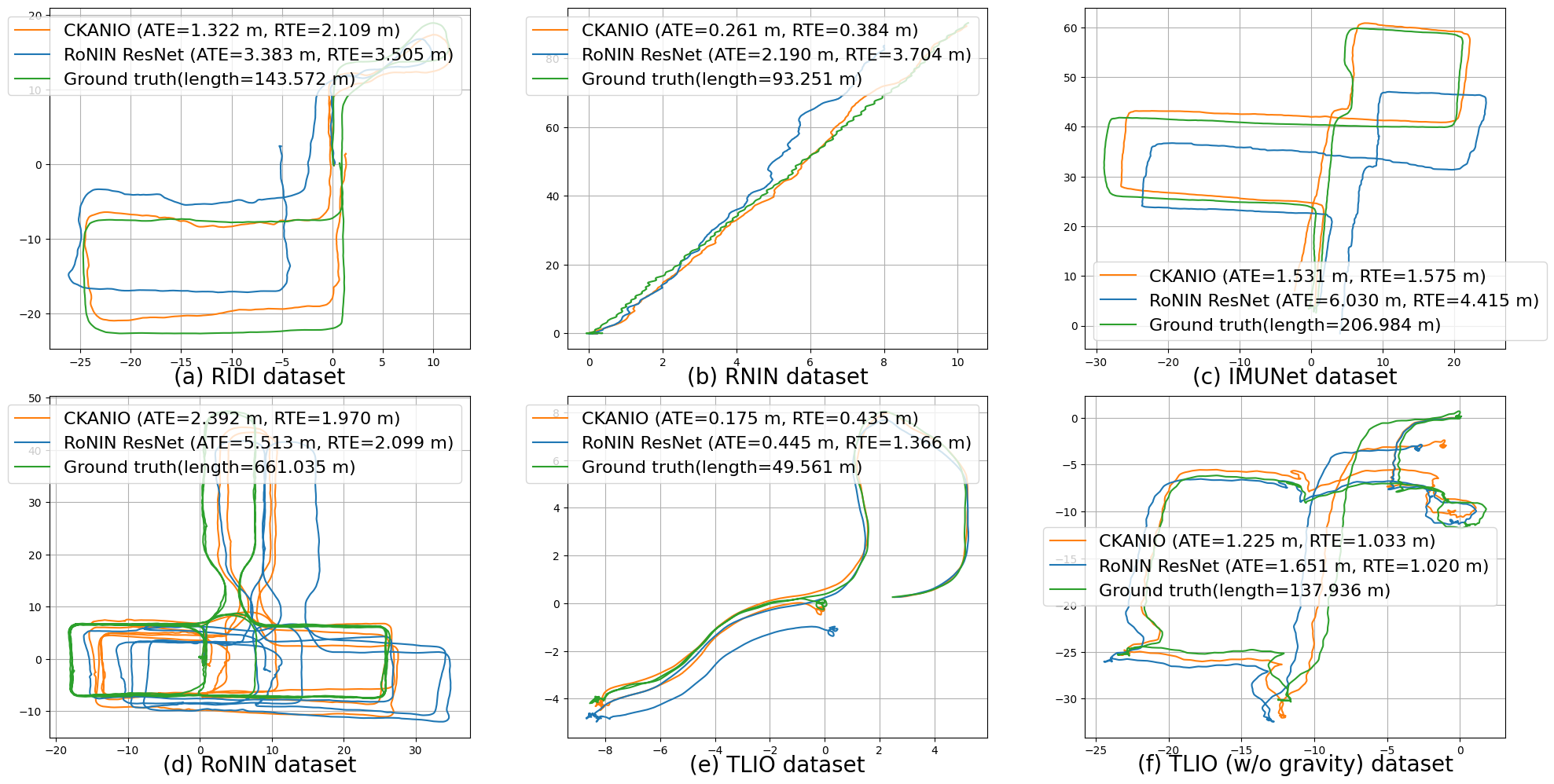}
\caption{Visual comparison of CKANIO and RoNIN ResNet on representative test trajectories from multiple datasets.}
\label{fig:visual_trajectory}
\vspace{-15pt} 
\end{figure*}

\subsection{Result Analysis}
\textbf{Quantitative Analysis.}
The left portion of Table~\ref{tab:cross_dataset_algorithm} reports the experimental results for all models across the five public datasets. CKANIO consistently outperforms the compared methods on most datasets, achieving lower ATE and RTE values. For example, on the RoNIN dataset and compared with the RoNIN ResNet, CKANIO reduces the ATE from 5.365m to 3.814m and the RTE from 3.390m to 3.272m. 

The right portion of Table~\ref{tab:cross_dataset_algorithm} also reports results obtained when models are trained on TLIO and on TLIO (w/o gravity). Removing gravity in the global-coordinate signals substantially reduces errors for nearly all evaluated algorithms. In a globally aligned coordinate frame, gravity acts effectively as a constant bias; removing this constant component allows the accelerometer measurements to better reflect the carrier's translational motion, which improves learning and numerical stability.

\textbf{Visualization Analysis.}
Fig.~\ref{RoNIN} shows the performance of CKANIO and three RoNIN variants on the RoNIN dataset (only RoNIN results are shown due to space constraints). As illustrated in Fig.~\ref{RoNIN}, CKANIO achieves the lowest PDE, underscoring its superior ability to converge precisely to the final position. This result confirms CKANIO's effectiveness in providing accurate and reliable positioning for inertial localization tasks.

Fig.~\ref{fig:visual_trajectory} illustrates representative test trajectories comparing CKANIO and RoNIN ResNet. In relatively simple scenarios (e.g., Fig.~\ref{fig:visual_trajectory}(b) and (e)), RoNIN ResNet predictions gradually deviate due to accumulated heading errors, whereas CKANIO demonstrates greater robustness to noise and more accurate trajectory reconstruction. In other cases (e.g., Fig.~\ref{fig:visual_trajectory}(a), (c), (d), and (f)), RoNIN ResNet exhibits noticeable drift after rotations and other nonlinear maneuvers; CKANIO better captures these dynamics and provides more accurate localization for long and complex trajectories.

\subsection{Ablation Study}
To validate the contribution of each component, we evaluate ablated variants of CKANIO, as shown at the bottom of Table~\ref{tab:cross_dataset_algorithm}. The CKANIO (w/o EKSA) variant (i.e., ResCKAN only) outperforms RoNIN ResNet, which confirms the effectiveness of the ResCKAN backbone. Incorporating EKSA to form the full CKANIO model further reduces ATE and RTE and enhances velocity prediction accuracy. These quantitative results indicate that both ResCKAN and EKSA contribute substantially to the overall performance gains.

\section{Conclusion}
We proposed CKANIO, which integrates Chebyshev KAN and kernel-based self-attention into IO to enhance the modeling of complex nonlinear motion features and contextual motion information. To the best of our knowledge, this is the first application of KAN within IO. CKANIO offers a new architectural reference for IO, and extending its evaluation to non-pedestrian IMU signals remains a promising direction for future research.

\section{Acknowledgements}
This work is supported by Science and Technology Major Program of Fujian Province (No. 2022HZ026007) and partly supported by the Science and Technology Planning Project of Fujian province (2022I0001).

\vfill\pagebreak

\bibliographystyle{IEEEbib}
\bibliography{refs}

\end{document}